\pdfoutput=1

\documentclass{esannV2}
\usepackage{graphicx}
\usepackage[latin1]{inputenc}
\usepackage{amssymb,amsmath,array}

\usepackage{datatool}

\usepackage{wrapfig}
\usepackage{hyperref}
\usepackage{url}
\usepackage{xspace}
\usepackage{stmaryrd}
\usepackage{float}
\usepackage{tabto}

\usepackage{algorithm2e}
\RestyleAlgo{ruled}
\SetAlgoNoLine
\SetAlgoNoEnd
\SetKwInput{KwInputs}{Inputs}
\SetKwInput{KwOutput}{Output}

\usepackage{cleveref}

\usepackage[numbers,sort&compress]{natbib}

\usepackage{glossaries}
\glsdisablehyper
\newglossaryentry{ch}{
  name={CH},
  description={Clever Hans},
  first={Clever Hans (CH)}
}
\newglossaryentry{ce}{
  name={CE},
  description={cross-entropy},
  first={cross-entropy (CE)}
}
\newglossaryentry{aga}{
  name={AGA},
  description={average group accuracy},
  first={average group accuracy (AGA)}
}
\newglossaryentry{xai}{
  name={XAI},
  description={Explainable AI},
  first={Explainable AI (XAI)}
}
\newglossaryentry{cfkd}{
  name={CFKD},
  description={Counterfactual Knowledge Distillation},
  first={Counterfactual Knowledge Distillation (CFKD)}
}
\newglossaryentry{tsne}{
  name={t-SNE},
  description={t-distributed Stochastic Neighbor Embedding},
  first={t-distributed Stochastic Neighbor Embedding (t-SNE)}
  plural={t-distributed Stochastic Neighbor Embeddings},
  firstplural={t-distributed Stochastic Neighbor Embeddings (t-SNE)}
}
\newglossaryentry{jtt}{
  name={JTT},
  description={Just Train Twice},
  first={Just Train Twice (JTT)}
}
\newglossaryentry{fl}{
  name={FL},
  description={Focal Loss},
  first={Focal Loss (FL)}
}
\newglossaryentry{dl}{
  name={DL},
  description={Deep Learning},
  first={Deep Learning (DL)}
}
\newglossaryentry{sota}{
  name={SOTA},
  description={state-of-the-art},
  first={state-of-the-art (SOTA)}
}
\newglossaryentry{oko}{
  name={OKO},
  description={Odd-$k$-Out},
  first={Odd-$k$-Out (OKO)}
}
\newglossaryentry{gan}{
  name={GAN},
  description={generative adversarial network},
  first={generative adversarial network (GAN)},
  plural={GANs},
  firstplural={generative adversarial networks (GANs)}
}
\newglossaryentry{dfr}{
  name={DFR},
  description={deep feature reweighting},
  first={deep feature reweighting (DFR)},
  plural={DFRs},
  firstplural={deep feature reweighting methods (DFRs)}
}
\newglossaryentry{sgd}{
  name={SGD},
  description={stochastic gradient descent},
  first={stochastic gradient descent (SGD)}
}
\newglossaryentry{bb}{
  name={BB},
  description={batch-balancing},
  first={batch-balancing (BB)}
}
\newglossaryentry{sce}{
  name={SCE},
  description={smooth counterfactual explorer},
  first={smooth counterfactual explorer}
  plural={SCE},
  firstplural={smooth counterfactual explorers (SCE)}
}
\newglossaryentry{erm}{
  name={ERM},
  description={empirical risk minimization},
  first={empirical risk minimization (ERM)}
}
\newglossaryentry{pde}{
  name={PDE},
  description={progressive data expansion},
  first={progressive data expansion (PDE)},
  plural={PDEs},
  firstplural={progressive data expansion techniques (PDEs)}
}

\newcommand{\D}{\ensuremath{\mathcal{D}}\xspace}

\newcommand{\celebsmile}{\text{C-Smile}\xspace}
\newcommand{\celebmale}{\text{C-Male}\xspace}
\newcommand{\camel}{\text{Camelyon17}\xspace}

\usepackage{multirow}
\usepackage{booktabs}
\usepackage{graphicx}
\usepackage{subcaption}
\usepackage{nicefrac}

\usepackage{tikz}
\usetikzlibrary{matrix,positioning,calc,backgrounds} 

\begin{document}
\title{Imbalanced Classification through the\\Lens of Spurious Correlations}

\author{Jakob Hackstein and Sidney Bender
%
\thanks{This work was supported by BASLEARN--TU Berlin/BASF Joint Laboratory, co-financed by TU Berlin and BASF SE.
}
%
\vspace{.3cm}\\
%
Machine Learning Group, TU Berlin, 10587 Berlin, Germany
%
}

\maketitle

\begin{abstract}
Class imbalance poses a fundamental challenge in machine learning, frequently 
leading to unreliable classification performance.
While prior methods focus on data- or loss-reweighting schemes, we view imbalance as a data condition that amplifies \gls{ch} effects by underspecification of minority classes.
In a counterfactual explanations-based approach, we propose to leverage Explainable AI to jointly identify and eliminate \gls{ch} effects emerging under imbalance.
Our method achieves competitive classification performance on three datasets and demonstrates how \gls{ch} effects emerge under imbalance, a perspective largely 
overlooked by existing approaches.
\end{abstract}
\section{Introduction}
\glsresetall

Classification under imbalance is a long-standing challenge in machine learning and remains highly relevant due to its prevalence in real-world applications.
Class imbalance destabilizes training, biases feature learning~\cite{francazi2023theoretical}, and causes overfitting to majority classes, altogether limiting classifier reliability.
Existing methods typically address these issues by implementing loss-reweighting schemes~\cite{lin2017focal} to emphasize learning minority classes.
While these approaches stabilize training, they leave a central challenge untouched: the minority class often 
provides insufficient data
to accurately model its underlying semantics, leaving it fundamentally underspecified.

We propose to view class imbalance through the lens of spurious correlations, arguing that insufficient information on minority classes encourages seemingly discriminative yet non-causal classification strategies. 
Given their inductive bias to favor simple features~\cite{shah2020pitfalls} and tendency to rely on spurious cues, classifiers are prone to adopt \gls{ch} solutions~\cite{lapuschkin2019unmasking}.
This perspective reveals a limitation in existing methods: they emphasize learning the correct \textit{classification outcome} but do not explicitly encourage a causal \textit{classification strategy}.
Inspired by this insight, we aim to address imbalance 
by mitigating \gls{ch} solutions that arise from minority classes. We study this approach in isolation by considering binary image classification tasks, which provide a controlled setting to analyze the interrelated effects of imbalance and spurious correlations.

\begin{figure}[t!]
\centering
 \includegraphics[width=0.95\textwidth]{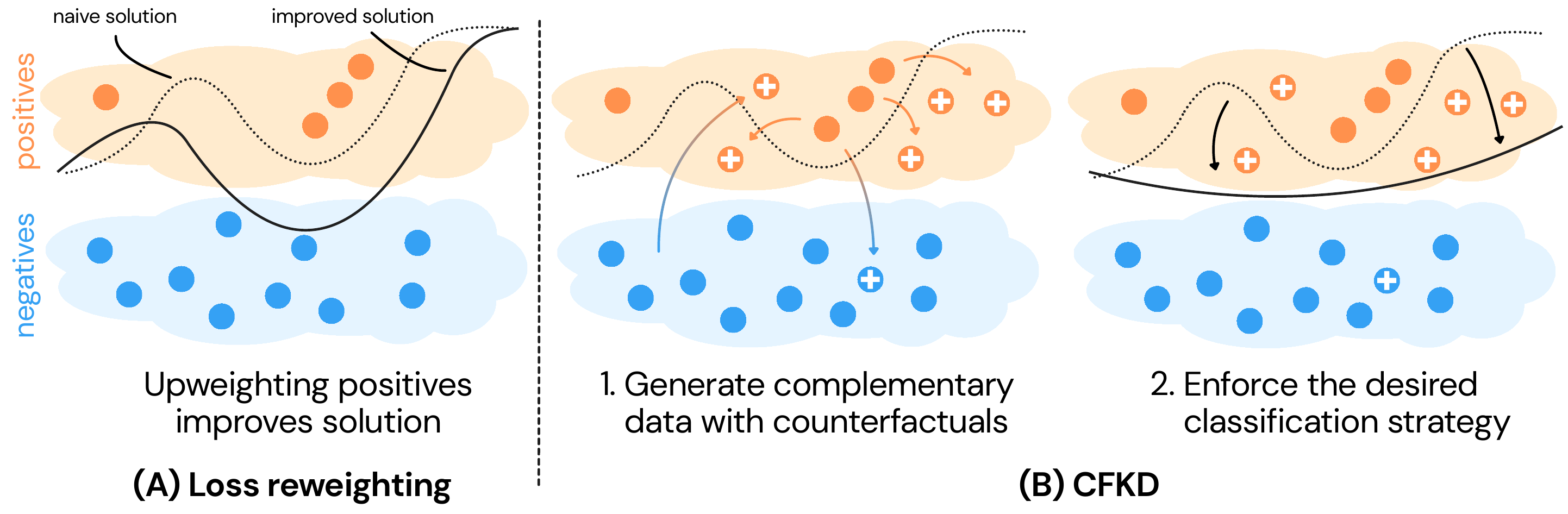}
\caption{
\textbf{(A)} Upweighting positives yields a better decision boundary, but \gls{ch} solutions persist.
\textbf{(B)} CFKD first generates complementary data using counterfactuals. All counterfactuals cross the initial decision boundary, but may not flip the true class label (false counterfactuals). Then, the desired classification strategy is distilled into the classifier,
explicitly prohibiting \gls{ch} solutions.
}
\label{fig:main_cfkd_figure}
\end{figure}

Enforcing causal behavior is challenging, as classification strategies are deeply entangled and categorical annotations often ambiguous. 
However, recent advances in \gls{xai} have introduced various methods to analyze classifier behavior~\cite{samek2021explaining,bach2015pixel,ribeiro2016should}.
Therefore, to access and influence behavior beyond mere outcome, we employ counterfactual explanations~\cite{dombrowski2024diffeomorphic, jeanneret2023adversarial}.
Counterfactuals are interpretable, minimally altered samples that suffice to flip a classifier's prediction, thereby exposing the decisive features behind its decision.
Our method builds on \gls{cfkd}~\cite{bender2023towards,bender2025mitigating}, 
which fine-tunes classifiers on domain expert-annotated counterfactuals. 
In contrast to reweighting schemes, this approach jointly uncovers \gls{ch} effects and effectively eliminates spurious correlations, as illustrated in~\Cref{fig:main_cfkd_figure}.
By extending the counterfactual-based \gls{cfkd} framework, we take a first step toward harnessing \gls{xai} to improve classification performance under imbalance and jointly ensure trustworthy employment in safety-critical settings.
Furthermore, we investigate the interplay between imbalance and spurious correlations through a series of experiments in both confounder-controlled and real-world settings.


\section{Theoretical Motivation}\label{sec:background}

To better understand how positive and negative samples corresponding to the minority and majority class, respectively, affect the emergence of spurious correlations, we examine a simplified example.
Consider a dataset with~$n$ positives and an imbalance ratio~$k \geq 1$, yielding $nk$ negatives. 
Each sample has two binary features, a causal $X_c$ aligned with class $Y$ by construction, and a spurious $X_s$ independent of class with $\Pr(X_s=1 \mid Y=y) = 0.5$ for $y \in \{0,1\}$.

We ascribe spurious correlations under imbalance to the underspecification of positives, as small~$n$ induces high variance in empirical estimates.
For~$n~=~1$,~$X_s$ appears perfectly predictive as its pattern matches the single positive sample. 
More generally, the probability that~$X_s$ aligns perfectly with all $n$ positive samples is~$2^{1-n}$, halving with every additional positive. Thus, with few positives, arbitrary features can mimic~$X_c$ and induce \gls{ch} solutions.
In contrast, negatives can refute the predictivity of non-causal features that small~$n$ alone might suggest.
For~$X_s$, the empirical negative prevalence follows~$p_s \sim \nicefrac{1}{nk}\,\mathrm{Bin}(nk, 0.5)$. As~$k$ increases,~$p_s$ concentrates around its true value $0.5$, revealing~$X_s$ as uninformative. Additionally, surplus negatives introduce sample variability and promote learning task-relevant features.

However, in practice, increasing $k$ quickly yields diminishing returns as severe imbalance causes practical issues during optimization and biases feature learning towards the majority class. Further,~$X_s$ may rarely occur in negatives (e.g., copyright tags~\cite{lapuschkin2019unmasking}), limiting $k$'s value in refuting \gls{ch} solutions.
Thus, while negatives assist in filtering futile features and refining representations, the trade-off between mitigating spurious correlations and avoiding severe imbalance allows \gls{ch} effects to persist.

\section{Method}\label{sec:method}
Mitigating \gls{ch} effects is a two-step process, as it involves (1) detecting reliance on spurious correlations and (2) subsequent removal of undesired feature reliance.
In our setting, designing a suitable method is challenging:
For (1), we must assume the emergence of multiple spurious correlations of arbitrary feature complexity.
These undesired feature dependencies are not known a priori and complicate (2), as we cannot rely on confounder labels during training.
Importantly, inherent ambiguity in the dataset limits the effectiveness of cost-sensitive techniques or data reweighting schemes, underscoring the need for a principled approach.

To account for these challenges, we utilize counterfactual explanations.
Formally, for a given image-class pair $(x,y)$ and a classifier $f$ with $f(x) = y$, a counterfactual $\tilde{x}$ is a semantically manipulated $x$ such that $f$ changes its prediction to a desired label $\tilde{y}$, i.e., $f(\tilde{x}) = \tilde{y}$.
A \textit{true counterfactual} flips the prediction by altering a causal feature, while a \textit{false counterfactual} alters a confounding feature, thereby revealing \gls{ch} effects.
Counterfactuals address (1) by detecting undesired feature reliance agnostic to the number and complexity of spurious features, and (2) by providing data to eliminate predictivity of spurious cues.

\subsection{Counterfactual Knowledge Distillation}
To effectively harness counterfactuals for imbalanced classification, we apply \gls{cfkd} as implemented in \Cref{alg:cfkd}.
\gls{cfkd} receives a base classifier $f$ trained on an imbalanced dataset $\D$, where $f$ is assumed to exhibit undesired behavior due to imbalance, along with a counterfactual explainer $\mathcal{E}$ and a teacher $\mathcal{T}$. In practice, $\mathcal{T}$ is a domain expert while our experiments rely on oracle models.
We draw a subset $\mathcal{S} \subseteq \D$ of image-class pairs and task $\mathcal{E}$ to generate counterfactual explanations $\smash{\tilde{\mathcal{S}}}$ according to the beliefs of $f$.
Since convincing $f$ to flip its prediction does not necessarily reflect ground-truth behavior, we query $\mathcal{T}$ to annotate each counterfactual $\smash{\tilde{x}}$ with a correct label $\tilde{y}^*$. If $\mathcal{T}$ also flips its prediction (i.e., $\smash{\tilde{y}^*} = \smash{\tilde{y}}$), the teacher agrees with the manipulated feature to cause a label flip and attests desired classifier behavior to $f$.
However, if the prediction of $\mathcal{T}$ remains (i.e., $\smash{\tilde{y}^*} = y$), then $f$ must base its classification behavior on a spurious correlation and a \gls{ch} solution is detected.
A refined classifier $f'$ is obtained by fine-tuning $f$ on $\D$ and $\smash{\tilde{\mathcal{S}}}$, which forces classification behavior to align with $\mathcal{T}$.

\gls{cfkd} is particularly suitable as it unifies the two-step process of (1) \gls{ch} detection and (2) mitigation.
Intuitively, \gls{cfkd} reveals classification behavior semantically using counterfactuals and leverages ground-truth annotations to provide explicit feedback on this behavior.
Thus, annotated counterfactuals serve as a proxy to correct the erroneous predictivity of arbitrary spurious correlations.

\begin{algorithm}[H]
\small
\caption{Counterfactual Knowledge Distillation}
\label{alg:cfkd}
\KwIn{%
  Classifier~$f$ trained on $\mathcal{D}=\{(x_i,y_i)\}_{i=1}^{N}$, explainer~$\mathcal{E}$, teacher~$\mathcal{T}$ \\
}
Draw subset $\mathcal{S} \subseteq \mathcal{D}$ to be explained by $\mathcal{E}$, $\smash{\tilde{\mathcal{S}}} = \emptyset$ \\
\For{$(x, y) \in \mathcal{S}$}{
        Generate counterfactual for $f$ with $\mathcal{E}$;\tabto{6.34cm}$\smash{\tilde{x}} \gets \mathcal{E}(f, x, 1-y)$
        
        Query $\mathcal{T}$ for true $\smash{\tilde{y}}^{*}$ and add to data;\tabto{6.2cm}$\smash{\tilde{y}}^* \gets \mathcal{T}(\smash{\tilde{x}}), \;\; \smash{\tilde{\mathcal{S}}}\texttt{.add}\big((\smash{\tilde{x}}, \smash{\tilde{y}}^*)\big)$
    }
\KwOut{
\text{Corrected classifier} $f' \gets f$\texttt{.finetune}($\mathcal{D} \cup \smash{\tilde{\mathcal{S}}})$}
\end{algorithm}


\section{Experiments}


We study \gls{cfkd}'s effectiveness in mitigating \gls{ch} effects emerging under imbalance on \camel~\cite{koh2021wilds}, \celebsmile, and \celebmale, where the latter two are variants of CelebA~\cite{liu2015faceattributes} binarized for \textit{Male} and \textit{Smiling}. 
Medical datasets provide safety-critical, confounder-rich settings, while CelebA is notoriously known for \gls{ch} effects.
To compose datasets, we adjust minority class size~$n$ and imbalance ratio~$k$ to create challenging dataset instances. For \camel, the hospital source site as a natural confounder is eliminated unless stated otherwise.

As a naive baseline, we train ImageNet-pretrained ResNet18 classifiers by optimizing \gls{ce} with L2 regularization and early-stopping.
For the base classifier to be corrected by \gls{cfkd}, we merely add \gls{bb} to prevent collapsed solutions. Advanced reweighting schemes are omitted intentionally to isolate correcting \gls{ch} effects.
We compare \gls{cfkd} to \gls{fl}~\cite{lin2017focal} due to its robustness and widespread use in imbalanced classification.
For \gls{cfkd}, we train oracle classifiers on well-conditioned datasets to substitute the domain expert during experiments. To correct classifiers, we follow the implementation of \gls{cfkd} and task \glspl{sce}~\cite{bender2025towards} with generating counterfactual explanations for 1000 samples.

\begin{figure}[b!]
\centering
\includegraphics[width=1.0\textwidth]{
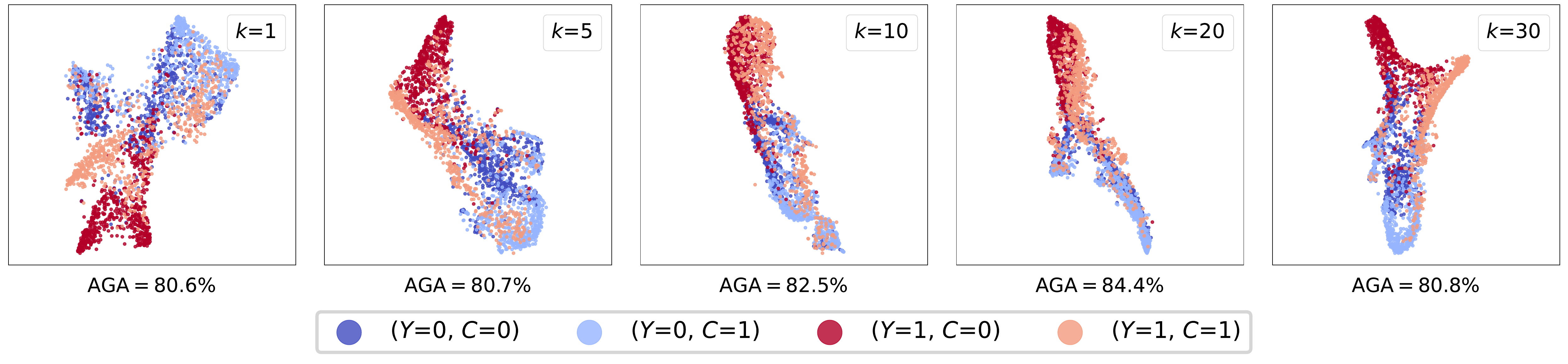
}
\caption{
AGA scores and t-SNE plots for \camel classifiers under moderate confounding~$C$ when increasing the imbalance ratio~$k$.
}
\label{fig:tsne}
\end{figure}

\subsection{Simulation on Confounder-Controlled Datasets}

\begin{table*}[b!]
\setlength{\tabcolsep}{2pt} %
\caption{
F1-Score (\%) for classification on \celebsmile, \celebmale, and \camel.
}
\label{tab:main_results_table}
\centering
\footnotesize
\begin{tabular}{c c | cccc | cccc | cccc}
\toprule
\multicolumn{2}{c|}{}
& \multicolumn{4}{c|}{\textbf{\celebsmile}} 
& \multicolumn{4}{c|}{\textbf{\celebmale}} 
& \multicolumn{4}{c}{\textbf{\camel}} \\ [1pt]
$n$ & $k$ 
& CE & CE+BB & FL & CFKD
& CE & CE+BB & FL & CFKD
& CE & CE+BB & FL & CFKD \\
\midrule
\multirow{3}{*}{100} 
 & 10 & 77.2 & 77.6 & 83.3 & \textbf{89.6} & 82.3 & 80.1 & 85.7 & \textbf{88.1} & 86.9 & 87.8 & 90.6 & \textbf{92.0} \\
 & 20 & 76.2 & 78.5 & 77.8 & \textbf{84.9} & 81.9 & 81.1 & 83.8 & \textbf{87.2} & 85.6 & 86.6 & 90.4 & \textbf{91.0} \\
 & 30 & 74.9 & 81.3 & 80.1 & \textbf{88.4} & 78.3 & 82.2 & 84.9 & \textbf{87.0} & 85.3 & 86.8 & 89.4 & \textbf{91.5} \\
\midrule
\multirow{3}{*}{200} 
 & 10 & 84.5 & 85.0 & 83.5 & \textbf{88.0} & 86.7 & 88.4 & 86.2 & \textbf{92.1} & 92.0 & 92.7 & 92.9 & \textbf{93.6} \\
 & 20 & 84.2 & 85.7 & 86.0 & \textbf{88.3} & 85.3 & 86.9 & 82.4 & \textbf{89.4} & \textbf{91.5} & 88.4 & 91.3 & 90.0 \\
 & 30 & 82.6 & 87.3 & 82.6 & \textbf{88.7} & 84.4 & 89.1 & 87.3 & \textbf{90.2} & 90.2 & \textbf{91.5} & 90.9 & \textbf{91.5} \\
\bottomrule
\end{tabular}
\end{table*}

In a preliminary experiment, we empirically complement our analysis from~\Cref{sec:background} by simulating a spurious correlation emerging under imbalance.
In particular, we study how surplus negatives influence classification performance as a trade-off between severe imbalance and refined features that might overcome \gls{ch} effects.
To this end, we utilize the hospital source site in \camel as a confounder $C$ and compose confounder-controlled datasets.
In the minority class,~$C$ co-occurs disproportionately with an observed prevalence of $90\%$ while it only occurs in $10\%$ of majority class samples. We set $n=100$ to deliberately underspecify the positive class and then ablate~$k$.
\Cref{fig:tsne} shows the resulting representations for all four $(Y, C)$ groups obtained by base classifiers (\gls{ce}+\gls{bb}) and corresponding \gls{aga}. 
The results agree with our analysis since,
for $k=1$, limited data yields weak features and the lowest \gls{aga}, as the classifier partly relies on $C$ (color saturation) rather than $Y$ (color temperature). For moderate $k$, $(Y,C)$ clusters become more distinct and \gls{aga} rises, indicating reduced \gls{ch} effects. However, for $k=30$, performance drops as severe imbalance adversely impacts training.
Thus, surplus negatives yield diminishing returns as refined representations cannot effectively overcome \gls{ch} effects.

\subsection{Results}

We present our main classification results in~\Cref{tab:main_results_table}.
As expected, the performance of naive \gls{ce} classifiers degrades as $k$ increases.
When \gls{bb} is employed, increasing~$k$ tends to improve performance slightly but yields diminishing returns, which agrees with our theoretical analysis and the previous trial.
Applying \gls{cfkd} to base classifiers substantially boosts performance across all datasets, and outperforms \gls{fl} in most cases. The results demonstrate that annotated counterfactuals successfully identify and eliminate spurious correlations, thereby improving both performance and reliability by preventing \gls{ch} solutions.

\section{Conclusion}

In this work, we address spurious correlations emerging under imbalance.
By applying the \gls{xai}-driven approach \gls{cfkd}, we successfully identify and mitigate the arising \gls{ch} effects leveraging teacher-annotated counterfactuals.
Thereby, we outperform popular baselines in imbalanced classification and jointly ensure trustworthy deployment in safety-critical environments. 
We hope this work inspires further analysis on the interrelated effects of imbalance and \gls{ch} solutions.




\begin{footnotesize}

\bibliographystyle{unsrt}
\bibliography{bibliography}

\end{footnotesize}


\end{document}